\documentclass{article}

    \PassOptionsToPackage{numbers, compress}{natbib}
    \usepackage[preprint]{style_2024}

\usepackage[utf8]{inputenc} % allow utf-8 input
\usepackage[T1]{fontenc}    % use 8-bit T1 fonts
\usepackage{hyperref}       % hyperlinks
\usepackage{url}            % simple URL typesetting
\usepackage{booktabs}       % professional-quality tables
\usepackage{amsfonts}       % blackboard math symbols
\usepackage{nicefrac}       % compact symbols for 1/2, etc.
\usepackage{microtype}      % microtypography
\usepackage{xcolor}         % colors
\usepackage{times}
\usepackage{latexsym}
\usepackage{epsfig}
\usepackage{graphicx}
\usepackage{amsmath}
\usepackage{amssymb}
\usepackage{algpseudocode}

\usepackage{multirow}
\usepackage{booktabs}
\usepackage{algorithm}
\usepackage{subfigure}
\usepackage{bbm}
\usepackage{colortbl}
\definecolor{mypink}{rgb}{.99,.91,.95}
\definecolor{mygreen}{rgb}{.9,.99,.9}
\definecolor{mygray}{gray}{.9}

\newcommand{\revision}[1]{{#1}}
\usepackage{xspace}
\newcommand{\maincoder}{R$^{2}$C$^{2}$-Coder\xspace}

\newcommand{\coder}{R$^2$C$^2$-Enhance\xspace}
\newcommand{\finetune}{R$^2$C$^2$-Enhanced tuning\xspace}
\newcommand{\Finetune}{R$^2$C$^2$-Enhanced Tuning\xspace}
\newcommand{\newcceval}{CrossCodeEval+\xspace}
\newcommand{\benchmark}{R$^2$C$^2$-Bench\xspace}
\usepackage{wrapfig}
\usepackage{arydshln}
\usepackage{makecell}
\usepackage{tcolorbox}
\usepackage{listings}
\usepackage{enumitem}

\usepackage{pifont} 
\title{R$^{2}$C$^{2}$-Coder: Enhancing and Benchmarking Real-world Repository-level Code Completion \\Abilities of Code Large Language Models}

\author{Ken Deng*, Jiaheng Liu*$^{\dagger}$, He Zhu*, 
Congnan Liu, Jingxin Li,  \\ \textbf{Jiakai Wang}, \textbf{Peng Zhao}, \textbf{Chenchen Zhang}, \textbf{Yanan Wu}, \textbf{Xueqiao Yin}, \\\textbf{Yuanxing Zhang},  \textbf{Zizheng Zhan}, \textbf{Wenbo Su}, \textbf{Bangyu Xiang}, \textbf{Tiezheng Ge}, \textbf{Bo Zheng} \\
       Alibaba Group \\
        \texttt{\{dengken.deng, ljh411989\}@taobao.com}}
\begin{document}

\maketitle
\let\oldthefootnote\thefootnote

\let\thefootnote\relax\footnotetext{* First three authors contributed equally.}
                \let\thefootnote\relax\footnotetext{$^\dagger$ Corresponding Author: Jiaheng Liu.}
\let\thefootnote\oldthefootnote

\begin{abstract}
% Code completion models have made significant progress in recent years.
Recently, repository-level code completion has drawn more attention in modern software development, and several baseline methods and benchmarks have been proposed.
However, existing repository-level code completion methods often fall short of fully using the extensive context of a project repository, such as the intricacies of relevant files and class hierarchies.
Besides,
the existing benchmarks usually focus on limited code completion scenarios,
which cannot reflect the repository-level code completion abilities well of existing methods.
To address these limitations, we propose the R$^2$C$^2$-Coder to enhance and benchmark the real-world repository-level code completion abilities of code Large Language Models,
% Specifically,
where
the R$^2$C$^2$-Coder includes a code prompt construction method R$^2$C$^2$-Enhance and a well-designed benchmark  R$^2$C$^2$-Bench.
Specifically,
first,
in R$^2$C$^2$-Enhance,
we first construct the candidate retrieval pool and then assemble the completion prompt by retrieving from the retrieval pool for each completion cursor position.
Second,
based on R$^2$C$^2$-Enhance,
we can construct a more challenging and diverse R$^2$C$^2$-Bench with training, validation and test splits,
where a context perturbation strategy is proposed to simulate the real-world repository-level code completion well.
% and meticulous filtering strategy is used for validation and test splits.
% better performance.
% Based on the code prompt construction strategy of  R$^2$C$^2$-Enhance,
% we can easily construct the R$^2$C$^2$-Bench, a more challenging and diverse Real-world Repository-level Code Completion benchmark.  
Extensive results on multiple benchmarks demonstrate the effectiveness of our R$^2$C$^2$-Coder.
\end{abstract}

% lack of structure information, the existing benchmark is not practical, to improve the robustness.
\section{Introduction}
Large Language Models for Code (Code LLMs), such as Codex \citep{chen2021evaluating}, CodeGen \citep{nijkamp2023codegen2,nijkamp2022codegen}, and StarCoder \citep{li2023starcoder}, have shown their power to enhance developer productivity as their promising results in code completion~\citep{athiwaratkun2022multi,austin2021program,chen2021evaluating,lu2021codexglue}. Besides, to evaluate these models in real-world scenarios, 
multiple code completion benchmarks are proposed.
% However, such an evaluation setting is over-simplified, and it is not able to reflect the model's capability in code completion accurately.
% Specifically, 
Recently,
the challenging repository-level code completion has drawn more attention,
where the extensive cross-file dependencies (i.e., contextual information from other code files within the same repository) are used to enhance the completion performance. 
Therefore,
a direct but complex solution is to enhance the context length of LLMs~\citep{position-interpolation,longlora,yarn}.
Nonetheless, increasing the context length significantly raises costs and is not affordable, especially concerning the large number of files in a given repository. 
Hence, several recent baseline methods~\citep{ding2023cceval,liu2023repobench} apply the widely-used Retrieval-Augmented Generation (RAG) strategy by retrieving the most relevant code snippets to improve the repository-level code completion.

\begin{wrapfigure}{r}{0.45\textwidth}
	\centering
 % \vspace{-2mm}
	\includegraphics[width=0.45\textwidth]{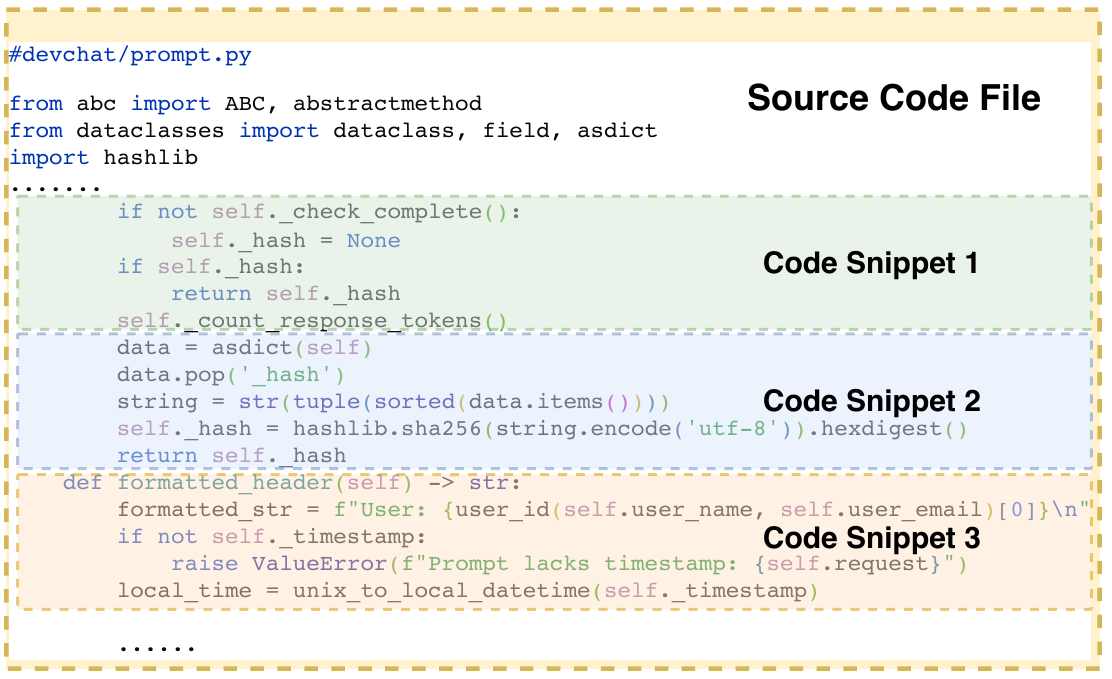}
 \vspace{-5mm}
	\caption{Examples of code snippets.}
	\label{fig:intro}
 \vspace{-5mm}
\end{wrapfigure}

However,
this vanilla strategy exhibits significant limitations.
% First, 
First,
it usually preprocesses the source code files into code snippets under a repository in Fig.~\ref{fig:intro}, and performs retrieval on these code fragments for retrieval query.
However, these fragments only preserve local and discrete information
and fail to cover more diverse and fundamental contexts of program semantics,
such as the intricate network of dependencies, shared utility functions, definition of the input and output parameters, etc.
% which indicates that existing methods often fail to consider the diverse contexts within the current repository. Note that the intricate network of dependencies, shared utility functions, inter-module method calls, definition of the input and output parameters and etc,
% are also fundamental to program semantics.
Besides,
% in Fig.~\ref{}(b),
as retrieved code snippets based on retrievers (e.g., BM25~\cite{robertson2009probabilistic}) usually contain irrelevant or noisy contexts,
% as the retriever (e.g., BM25~\cite{} or 
% Jaccard~\footnote{}
% Jaccard~\cite{}) cannot ensure,
which may affect the completion results,
it is important to evaluate and improve the robustness of the noisy contexts for Code LLMs. 
Moreover,
we observe that the existing repository-level code completion benchmarks focus on limited completion scenes and cannot fully simulate real-world usage well (See Section~\ref{sec:compare}).
% \begin{figure}[t]
%     \centering
%     \includegraphics[width=0.6\linewidth]{Styles/coder-intro-final.pdf}
%     \caption{Illustration of different contexts.}
%     \label{fig:intro}
% \end{figure} 

% fails to consider the global structure information (e.g., cross-file dependencies, inter-module method calls).
% Besides,
% the current methods 

To address these limitations,
we propose the \textbf{\maincoder} to enhance and benchmark the Real-world Repository-level Code Completion abilities of Code LLMs,
where a code prompt construction method (i.e., \textbf{\coder}) and a challenging benchmark  (i.e., \textbf{\benchmark}) are proposed.
% Specifically,
% for enhancing,
% First,

First, \coder includes two stages (i.e., candidate retrieval pool construction, and completion prompt construction).
% in Fig.~\ref{}(c),
% we propse
% extract both global and local contexts to improve the repository-level code completion.
% Specifically,
% For understanding,
% Specifically,
In the retrieval pool construction stage,
we first build the retrieval pool using the abstraction context based on the parser generator tool (e.g., Tree-sitter\footnote{\url{https://tree-sitter.github.io/tree-sitter/}}) and the snippet context, where the abstraction context aims to represent the coarse-grained global information of each programming file and the snippet context is built by extracted code fragments to
% the intra-snippet context and the inter-snippet context to 
provide the fine-grained local information for completion. 
Then,
in completion prompt construction stage,
% Then, 
for the current cursor position\footnote{The cursor position denotes the location, where code completion is about to be triggered.},
we first build the retrieval query and produce the retrieved contexts by performing the retrieval between the retrieval query and the retrieval pool,
where the retrieved contexts are then combined with the current code file to generate the completion prompt.
Finally,
% which is then 
% where the retrieval prompt is
the completion prompt is sent to Code LLMs to generate the completion response.

% enhance the robustness of the trained LLMs in constructing the retrieval prompt for training, where the context perturbation is not used at inference.

Second,
based on \coder,
we can easily build a real-world repository-level code completion benchmark called \textbf{R$^2$C$^2$-Bench},
which includes the training, validation and test splits in 4 languages.
Note that when producing \benchmark,
% in \coder,
we also apply a context perturbation strategy to simulate more diverse and challenging code completion samples for generating the completion prompt.

By performing experiments on the validation and test splits of \benchmark,
we observe that our  \coder achieves significant improvements when compared with the existing methods without training.
Besides,
when we further fine-tune the Code LLMs using the training split of \benchmark,
better results are obtained on the validation and testing splits of \benchmark.
Moreover,
we also validate the generalization abilities of \coder on the widely-used CrossCodeEval benchmark.

The contributions of our \maincoder are  as follows:
\begin{itemize}
    \item We investigate the limitations of existing repository-level code completion (e.g., lack of sufficient context and benchmark with limited scenarios) and propose the \maincoder including \coder and \benchmark to enhance and benchmark real-world repository-level code completion abilities of Code LLMs.
    % analyze the influence of the existing face distillation method from the distribution of positive pairs and negative pairs. The former represents intra-class compactness, and the latter represents the distribution between classes.
    \item  For \coder, we propose to construct the candidate retrieval pool with abstract and snippet contexts and generate the completion prompt using context retrieval and prompt assemble.
    % Besides,
    % in training,
    % we also apply a simple context perturbation strategy to further enhance the robustness of Code LLMs.
    % by randomly replacing the relevant context with the irrelevant context in training.
    Based on \coder,
    we build a new repository-level code completion benchmark called \benchmark with training, validation, and testing splits,
    where a context perturbation strategy is used to simulate the real-world completion scenes better.
    % sets are xx, xx, and xx,
% respectively.
    % \item In our E$^{2}$-LLM, we first provide the initial version (\textbf{API-Ini}) by randomly selecting the scale parameter proposed in position interpolation and then introduce the better version (\textbf{API-Pro}) by randomly changing the starting position index of RoPE to make the LLMs more robust to the relative differences and ignore the absolute position index values.
    \item  
Comprehensive experimental results on multiple benchmark datasets demonstrate the effectiveness and efficiency of our \maincoder.
% \item 
% % Besides,
% Our method also preserves the short-context understanding ability well when compared to existing long-context extension methods,
% where short-context understanding ability usually degrades for existing methods.
\end{itemize}

\section{Related Works}
\noindent\textbf{Code Large Language Models.}
The successive achievements of generative language modeling \citep{black2022gptneox, gpt-neo, brown2020gpt3, radford2019gpt2, touvron2023llama, gpt-j,mtbench,concepthmath,liu20242,unicoder,compress,wang2023rolellm,guo2023owl,zhang2024mapneo,du2024chinese} have inspired extensive studies on generative AI for software engineering.  While closed-source models \citep{achiam2023gpt4, chen2021evaluating, chowdhery2023palm} achieve significant dominance in the benchmark leaderboards \citep{chen2021evaluating, hendrycks2020measuring, liu2023evalplus}, their inaccessible model checkpoints and source code are detrimental to subsequent innovation. In contrast, plenty of open-source models, for instance, Code T5 \citep{wang2023codet5+, wang-etal-2021-codet5}, CodeGen \citep{nijkamp2023codegen2, nijkamp2022codegen}, StarCoder \citep{allal2023santacoder, li2023starcoder, lozhkov2024starcoder2}, Code Llama \citep{roziere2023codellama} and DeepSeekCoder \citep{guo2024deepseek}, have been proposed and have substantially driven the development of code intelligence.
% Although huge efforts have been paid to improve the general performance, the model alone is not sufficient to cope with some real-world scenarios, such as repository-level code completion \citep{liu2023repobench, ding2023cceval, jimenez2023swe, li2024evocodebench}.

\noindent\textbf{Repository-level Code Completion.}
% Repository-level code completion focuses on filtering those code snippets from the code base that are most relevant to the location  to be completed.
Existing methods are usually based on a retrieval augmented generation (RAG),
and we categorize related works into ranking-based and fusion-based methods. Ranking-based approaches \citep{liao2023a3codgen, pei2023better, phan2024repohyper, zhang2023repocoder} explicitly select those code snippets in the repository that are highly similar to the incomplete code.
% where the similarity measure can be based on lexical retriever (e.g. BM25 \citep{robertson2009probabilistic}) or dense retriever (e.g. UniXCoder \citep{guo2022unixcoder}). 
In contrast, fusion-based methods \citep{ding2022cocomic, shrivastava2023repofusion, shrivastava2022repository} {focus on organizing the repository-level context to be jointly modeled with the language model and allow the model to automatically select the most relevant information.
% related to the unfilled loci}. 
% Apart from the aforementioned studies, a few recent works focus on refining the generation process \citep{bairi2023codeplan, agrawal2023guiding} to improve the executable rate of code completion.}
% TODO: 锐评

\noindent\textbf{Code Benchmark.}
Program synthesis is to prompt code models to solve programming problems using the input description. The generated code snippet should be functional to pass all test cases~\citep{athiwaratkun2022multi, austin2021program, chen2021evaluating, gu2024cruxeval, lai2023ds1000, liu2023evalplus, yu2024codereval}. 
% \citet{allal2023santacoder} propose a single-line-infilling benchmark using the exact match as the metric. 
% With the PAL framework \citep{gao2023pal}, benchmarks originally designed for testing the mathematical reasoning capabilities of language models can also be adapted to code models. \citep{cobbe2021gsm8k, hendrycks2020measuring, gao2023pal, patel2021nlp, lu2022dynamic, miao2020diverse, gou2023tora}. 
Besides, numerous benchmarks can be used to comprehensively evaluate code models, such as code translation~\citep{jiao2023evaluation, yan2023codetransocean, zhu2022xlcost}, code retrieval~\citep{huang2021cosqa, husain2019codesearchnet, li2024procqa, lu2021codexglue}, and vulnerability repair~\citep{huq2022review4repair, prenner2023runbugrun, richter2022tssb, tian2024debugbench}. Recently, several benchmarks have been proposed on repository-level code completion~\citep{agrawal2023guiding, allal2023santacoder, bairi2023codeplan, ding2022cocomic, liu2023repobench, pei2023better, shrivastava2022repository, shrivastava2023repofusion, zhang2023repocoder}.
% However, most of these benchmarks collect fewer than one hundred repositories and study only one programming language. To comprehensively evaluate a code auto-completion system, 
For example, RepoBench \citep{liu2023repobench} measures the retrieval and completion performance of the system. CrossCodeEval \citep{ding2023cceval} evaluates three cross-file prompting levels: in-file, retrieval, and retrieval within references. However, RepoBench only supports two programming languages and neglects the FIM setting. CrossCodeEval only considers explicit file imports, resulting in few cross-file dependencies and low multi-line prediction difficulty.

\section{\maincoder}
% \subsection{Preliminary on Repository-level Code Completion}

% \subsection{HierCoder}
In this section, we discuss the details of \textbf{\maincoder},
which includes a code prompt construction method \textbf{\coder} and a new benchmark \textbf{\benchmark}.
For \coder,
% specifically,
in Fig.~\ref{fig:overview},
we first build the candidate retrieval pool based on the abstract context and the snippet context to produce the coarse-grained global information of each programming file and the fine-grained local information of each code fragment,
respectively.
% or completion. 
Then, 
for the current triggered cursor location,
we produce the retrieval query and obtain the retrieved contexts based on the completion query and the candidate context pool,
where the retrieved contexts and the current in-file context are used to
construct the completion prompt.
% For \benchmark,
Based on \coder,
we can easily produce the \benchmark,
where a context perturbation strategy is used to generate more diverse and challenging completion scenes.
% to enhance the robustness of the trained LLMs in constructing the retrieval prompt for training, where the context perturbation is not used at inference.

% by  chunked code as the additional structure information and then obtain the similarities between the completion context and the context.
\begin{figure}[t]
    \centering
    \includegraphics[width=0.98\linewidth]{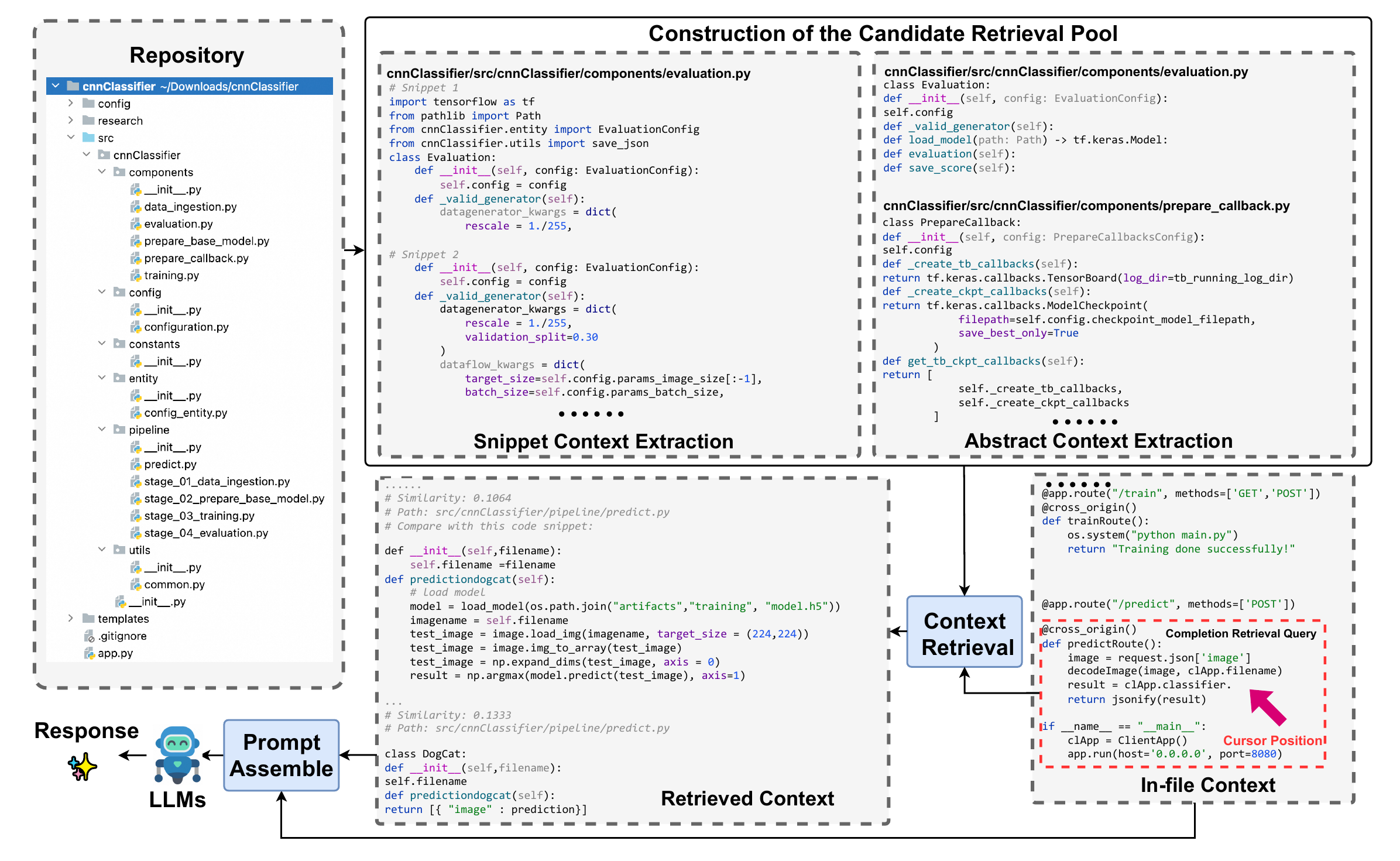}
    % \vspace{-4mm}
    \caption{Overview of our \coder. For the current completion cursor position, we first generate the retrieval query using the prefix and suffix contexts. Then, we perform context retrieval between the retrieval query and the pre-constructed candidate retrieval pool to produce the retrieved contexts. After that, we use the in-file context of the current code and the retrieved contexts to assemble the completion prompt,
    which is then sent to LLMs to generate the completion response. }
    % \vspace{-5mm}
    \label{fig:overview}
\end{figure} 
\subsection{R$^2$C$^2$-Enhance}
\label{sec:enhance}
\subsubsection{Construction of the Candidate Retrieval Pool}
\label{sec:candidate_pool}
Given a repository, we extract abstract and snippet contexts to construct the candidate retrieval pool.
% based on  the extracted abstraction context and snippet context.

% \begin{figure}[t]
%     \centering
%     \includegraphics[width=0.6\linewidth]{Styles/coder-ast.pdf}
%     \caption{An example of the abstract syntax tree generated by the Tree-sitter tool.}
%     \label{fig:ast}
% \end{figure} 

\begin{wrapfigure}{r}{0.5\textwidth}
	\centering
 \vspace{-5mm}
	\includegraphics[width=0.5\textwidth]{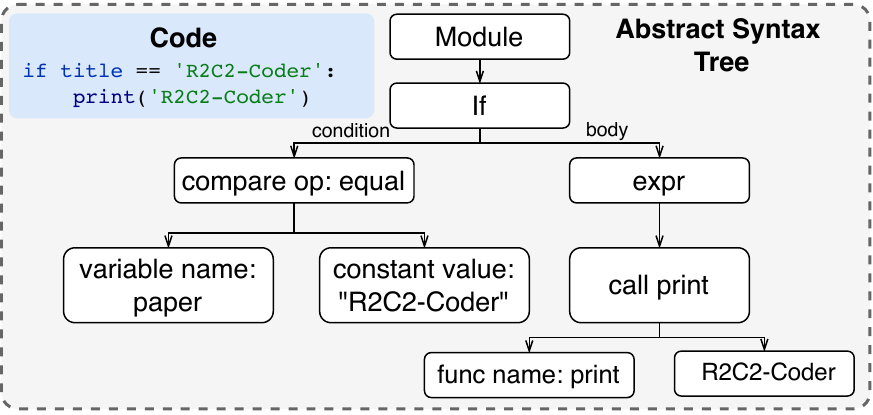}
 \vspace{-5mm}
	\caption{An example of the abstract syntax tree generated by the Tree-sitter tool.}
	\label{fig:ast}
 \vspace{-3mm}
\end{wrapfigure}
\noindent\textbf{Abstract Context Extraction.}
In Fig.~\ref{fig:overview},
we propose to use the parser generator tool (i.e., Tree-sitter\footnote{\url{https://tree-sitter.github.io/tree-sitter/}}) to extract the abstract context for each file in the repository. 
Specifically,
we use Tree-sitter to generate the abstract syntax tree in Fig.~\ref{fig:ast},
% which consists of several nodes
and then preserve these nodes to represent the declarations (e.g., functions, variables, classes etc.).
After that,
for each node,
we remove the redundant information (e.g., middle variables, empty lines, comments etc.) and only preserve the information related to the definition for each declaration.
Note that the extraction process is easy and efficient (See Appendix~\ref{app:ast_sec} for more details.).
Moreover, in Fig.~\ref{fig:abs} of  Appendix~\ref{app:ast_sec},
we provide examples of the extracted abstract contexts and observe that the abstract context mainly provides the coarse-grained global information for each code file.
% For example, in Fig.~\ref{}(a), the abstract context of function xx has the function name, input and output parameters. 

\noindent\textbf{Snippet Context Extraction.}
For the snippet context,
% as shown in Fig.~\ref{},
% we propose to extract the intra-file and the inter-snippet contexts as follows. 
% For the intra-snippet context,
% we directly follow~\cite{ding2023cceval} to maintain the current file, where the completion query belongs.
% propose to 
% For the inter-snippet context,
we iteratively scan the files in the repository and extract contiguous $M$ lines (in all our experiments, we set $M=10$ follow~\cite{ding2023cceval}) of overlapping code fragments, which are the candidates for context retrieval.

\subsubsection{Construction of the Completion Prompt}
\label{sec:prompt}
\noindent\textbf{Context Retrieval.}
In  Fig.~\ref{fig:overview},
for the current cursor position,
we first build the retrieval query based on the prefix and suffix contexts.
% which aims to find the relevant context to construct the retrieval prompt.
Specifically,
for the current cursor position,
we first use the previous $P$ lines and subsequent $S$ lines as the prefix and suffix contexts, respectively. Therefore,
the retrieval query usually has $P+S$ lines.
Note that if the prefix and suffix have one overlapped line, the retrieval query has $P+S-1$ lines.
Then,
% last $N$ lines (we set $N=10$) of the in-snippet context.
we use the retriever (e.g., BM 25~\citep{robertson2009probabilistic}) to calculate the similarity between the retrieval query and the candidate retrieval pool.
After that,
we can select the retrieved contexts based on the similarities.
Note that in the retrieval process,
we remove the abstract context and snippet context of the current code file,
where the cursor position is located.

\noindent\textbf{Prompt Assemble.}
We assemble the completion prompt by using the current code file, and retrieved contexts.
% where the retrieved contexts include abstract context and snippet context.
% completion prompt,
Specifically,
% After that,
% for the co
% We retain all information about the code file where the current completion site is located
we first follow~\cite{ding2023cceval} to maintain all context of the current code file as the in-file context, where the completion cursor position is located.
Then,
we preserve the top-$K$ similar abstract contexts as the coarse-grained global structure information,
% which provide the definitions of each functi
After that,
we append the most relevant snippet context based on the similarity scores until the maximum number of tokens is $N$,
where we set $N$ as 4,096 by default.
Finally,
we can obtain the assembled prompt with $N$ tokens,
which will be sent to the Code LLMs for generating the completion results.

% After that,
% we can assemble the newly augmented completion prompt for these $\%P$ training samples,
% In this way,
% we can obtain  the retrieved contexts
% which are not very relevant to the completion query.
% Finally,
% we seed the augmented completion prompt to the Code LLMs, and generate the completion results in training.

% and use the top-5 similar code snippets as the cross-snippet context, see ``Retrieval Context" in Figure~\ref{fig:prompt_format}.
% We consider a maximum of 512 BPE tokens for such context, and the rest of the tokens will be truncated. Figure~\ref{fig:prompt_format} illustrates\footnote{For the convenience of illustration, we remove the irrelevant code snippets from the retrieved context to avoid confusion. In practice, the retrieved context spans across a fixed length of lines, which includes useful information as well as its surrounding lines. More implementation details are in Appendix~\ref{subsec:appendix_cfc_retrieval}.} the retrieved context and the corresponding prompt for the model to complete. Given the in-snippet context as a query, the RG framework successfully retrieves the class definition of \texttt{CaseConverter} that is in another file for utilities. We further wrap the class definition into a template as code comment and use it as the cross-snippet context. To build the retrieval prompt, we prepend the retrieved context to the in-snippet context.

% \subsection{Training and Inference}

\subsection{\benchmark}
In this section,
we provide the details of our \benchmark with training, validation and test splits.
% respectively.

\subsubsection{Dataset Generation}
\label{sec:dataset_generation}
\noindent\textbf{Initial Repository Collection.}
We collect permissively licensed repositories from GitHub. 
% To mitigate potential data leakage issues,\footnote{Overlapping between R$^2$-Coder and data used to pretrain code LMs.} we focus on repos that were created \textit{recently} and not forks. 
Specifically, 
% we collected repos created between 2023-03-05 to 2023-06-15 on 2023-09-01. The time span ensures sufficient data collected with no overlap with the training data of many existing code LMs released before mid-2023, no matter whether the data is publicly available or not. 
{following~\citep{ding2023cceval},
we contain four languages and keep only repositories with the number of stars $>=3$.
Meanwhile,
% Then we filter out repositories with fewer than 10 or more than 50 source code files. Finally, we remove these repositories with at least one source code file that exactly matches one of the code files in the Stack \citep{kocetkov2022stack} dataset.
% Meanwhile,
% Next, 
we exclude repositories that contain fewer than 10 or more than 50 source code files,
and we eliminate any repositories that have at least one source code file identical to any file found in the Stack dataset \citep{kocetkov2022stack}.
As a result, we ended up with 54972, 51796, 49790, and 35410 repositories for Python, Java, TypeScript, and C\#, respectively.}

\noindent\textbf{Completion Cursor Position Generation.}
\label{sec:data_generation}
Based on \coder,
% \noindent
given completion cursor position,
we can easily generate the completion prompt.
Therefore,
it is critical to generate informative and sufficient cursor positions for our \benchmark.
Specifically,
{in Fig.~\ref{fig:ast},
for each file in the repository,
we reuse the abstract syntax tree generated in the abstract context extraction stage,
and randomly choose one node of the tree as the cursor position to be triggered for completion.
Then,
for the selected node,
we can easily generate one completion prompt sample for our \benchmark.
% we use the prefix and suffix nodes as the 
}
% In this way,
% we can generate

\begin{wraptable}{r}{80mm}
% \vspace{-8pt}
\centering
\def\arraystretch{1.0}%
\resizebox{1\linewidth}{!}
{
\begin{tabular}{l r r r r}
\toprule
Feature & Python & Java & TypeScript & C\#\\
\midrule
\# Repositories & 260 & 397 & 451 & 245 \\
\# Files & 2500 & 2500 & 2500 & 2500 \\
\# Examples & 4394 & 6900 & 6506 & 5028 \\
Avg. \# cross files in prompt &6.4&7.8&6.8&10.1\\
Avg. \# lines in prompt & 320.5 & 333.7 & 365.7 & 373.8 \\
Avg. \# tokens in prompt & 3192.25 & 3262.3 & 3200.52& 3313.23 \\
Avg. \# lines in reference & 1.73 &1.75 & 2.02 & 1.78 \\
Avg. \# tokens in reference & 17.47 & 14.22 & 15.00 & 13.32 \\
\bottomrule
\end{tabular}
}
\caption{The statistics of the testing split of \benchmark.} 
\label{tab:data_stat}
\vspace{-2mm}
\end{wraptable}

\noindent\textbf{Context Perturbation.}
In real-world repository-level code completion scenes,
we cannot always retrieve the relevant context well.
For example, in certain software development scenes, we create code files to implement different usages under a repository.
However,
the relationships between these code files are relatively low,
which indicates that the constructed completion prompt contains many irrelevant contexts.
Therefore, to simulate the real-world repository-level code completion scene and improve the semantic variety of the completion prompt,
we introduce a so-called context perturbation strategy.
Specifically,
for $Q\%$ completion cursor position (we set $Q=10$ by default),
we first sort the contexts based on the similarities and follow the uniform distribution to determine the proportion $R\%$. Then we randomly discard top $R\%$ contexts with high similarities.
% where we follow the uniform distribution to determine the proportion
After that,
we perform the context retrieval based on the retrieval query and the remaining contexts,
and the generated retrieved contexts are then combined with the current file to produce the completion prompt as discussed in Sec.~\ref{sec:prompt}.

Based on the above meticulous procedure,
we can produce sufficient  completion prompt samples using these repositories, and randomly select 400,000 as training samples,
where each language has 100,000 samples. 
For the validation and testing samples,
% After that,
% \jh{After that,
to ensure the quality of the dataset,
we apply a series of rule-based and model-based post-processing filter strategies discussed in  Appendix~\ref{app:quality}.
% and we obtain xx samples. 
% Finally,
 % xx, xx as the training, validation and test samples.

% \subsubsection{Post-processing and Quality Control}
% \label{sec:data_quality}
% See Appendix  \ref{section:human_annotation} for more details.

\subsubsection{Dataset Statistics}
\label{sec:data_stats}

% \begin{table}
% % \vspace{-8pt}
% \centering
% % \def\arraystretch{1.0}%
% % \resizebox{1\linewidth}{!}
% % {
% \begin{tabular}{l r r r r}
% \toprule
% Feature & Python & Java & TypeScript & C\#\\
% \midrule
% \# Repositories & 260 & 397 & 451 & 245 \\
% \# Files & 2500 & 2500 & 2500 & 2500 \\
% \# Examples & 4394 & 6900 & 6506 & 5028 \\
% Avg. \# cross files in prompt &6.4&7.8&6.8&10.1\\
% Avg. \# lines in prompt & 320.5 & 333.7 & 365.7 & 373.8 \\
% Avg. \# tokens in prompt & 3192.25 & 3262.3 & 3200.52& 3313.23 \\
% Avg. \# lines in reference & 1.73 &1.75 & 2.02 & 1.78 \\
% Avg. \# tokens in reference & 17.47 & 14.22 & 15.00 & 13.32 \\
% \bottomrule
% \end{tabular}
% \caption{The statistics of the testing split of \benchmark. } 
% \label{tab:data_stat}
% % \vspace{-2mm}
% \end{table}

% \paragraph{Statistics} 
The statistics of \benchmark are shown in Table \ref{tab:data_stat}\footnote{We use the StarCoder tokenizer \citep{li2023starcoder} to compute the number of tokens.}. We have provided the statistics of CrossCodeEval and validation split of \benchmark in Appendix~\ref{app:cceval}

\begin{table*}[!th]
	\centering
	\resizebox{0.98\textwidth}{!}{
		\begin{tabular}{l c c c c c c c c c c}
			\toprule
			\multirow{2}{*}{\bf Model}  &
			\multicolumn{2}{c}{\textbf{Python }}&
			\multicolumn{2}{c}{\textbf{Java }}&
			\multicolumn{2}{c}{\textbf{TypeScript }}&
			\multicolumn{2}{c}{\textbf{C\# }}&
			\multicolumn{2}{c}{\textbf{Average}}\\
			\cmidrule(lr){2-3} \cmidrule(lr){4-5} \cmidrule(lr){6-7} \cmidrule(lr){8-9}\cmidrule(lr){10-11}
			&{EM} &{ES} & {EM} & {ES} & {EM} &{ES} & {EM} & {ES} &{EM} &{ES} \\\midrule
			
			\revision{StarCoder-7B} & 
			17.0 & 49.5 & 21.6 & 55.9 & 18.1 & 53.6 & 17.6 & 51.0 & 18.6 & 52.5  \\
			\hdashline\noalign{\vskip 0.4ex}\; + \revision{\coder}&
			21.4 & 46.2 & 27.6 & 51.9 & 24.8 & 52.8 & 25.2 & 45.8 & 24.8 & 49.2 \\
			\hdashline\noalign{\vskip 0.4ex}\; + \revision{\coder w.o. Abs.}&
			21.5 & 46.2 & 26.1 & 51.4 & 24.6 & 52.8 & 24.9 & 45.7 & 24.3 & 49.0 \\
			\hdashline\noalign{\vskip 0.4ex}\; + \revision{\Finetune}& 
			34.0 & 65.0 & 45.3 & 74.4 & 37.0 & 68.9 & 45.0 & 71.6 & 40.3 & 70.0 \\
			\midrule
			\revision{Code Llama-7B} & 
			13.7 & 44.1 & 20.6 & 54.5 & 17.4 & 51.2 & 16.0 & 48.4 & 16.9 & 49.5  \\
			\hdashline\noalign{\vskip 0.4ex}\; + \revision{\coder} & 
			16.3 & 39.0 & 22.3 & 45.0 & 22.1 & 46.5 & 18.8 & 36.0 & 19.9 & 41.6  \\
			\hdashline\noalign{\vskip 0.4ex}\; + \revision{\coder w.o. Abs.} &
			16.2 & 38.9 & 21.6 & 44.8 & 22.0 & 46.4 & 19.1 & 36.9 & 19.7 & 41.7 \\
			\hdashline\noalign{\vskip 0.4ex}\; + \revision{\Finetune}& 
			34.5 & 66.5 & 47.4 & 77.2 & 40.2 & 71.7 & 48.0 & 74.7 & 42.5 & 72.5 \\
			\midrule
			\revision{DeepSeekCoder-6.7B} & 
			19.4 & 52.5 & 24.2 & 59.3 & 22.0 & 58.8 & 20.7 & 54.1 & 21.6 & 56.2 \\
			\hdashline\noalign{\vskip 0.4ex}\; + \revision{\coder} &
			25.8 & 51.3 & 31.8 & 56.5 & 29.7 & 57.8 & 33.1 & 53.6 & 30.1 & 54.8  \\
			\hdashline\noalign{\vskip 0.4ex}\; + \revision{\coder w.o. Abs.} &
			25.8 & 51.3 & 30.7 & 55.6 & 29.6 & 58.0 & 31.9 & 52.5 & 29.5 & 54.3 \\
			\hdashline\noalign{\vskip 0.4ex}\; + \revision{\Finetune}& 
			38.0 & 69.6 & 49.2 & 78.0 & 43.0 & 73.8 & 52.3 & 77.5 & 45.6 & 74.7 \\
			\bottomrule
		\end{tabular}
	}
	\caption{ 
		Exact match (\%) and edit similarity (\%) performance on \benchmark. The baseline results are produced under the ``In-file Context Only'' setting while ``+\coder'' means the input is prepended with retrieval results fetched by \coder at inference. In ``+ \coder w.o. Abs.'', we report results without using abstract contexts in the candidate retrieval pool. ``\coder w/ F.T.'' means we fine-tune these LLMs on the training set of \benchmark with \coder.} 
	\label{tab:main_benchmark}
\end{table*}
\section{Experiments}
\label{sec:experiments}
% \subsection{Code LM baselines}
\label{subsec:models}
We conduct experiments on CrossCodeEval \citep{ding2023cceval}, \newcceval, and \benchmark with three popular Code LLMs (i.e., \textbf{StarCoder-7B}~\citep{li2023starcoder}, \textbf{DeepSeekCoder-7B} \citep{guo2024deepseek} and \textbf{Code Llama-7B} \citep{roziere2023codellama}) (See Appendix~\ref{app:models} for more details).
For \newcceval, we collect the original repositories of CrossCodeEval \citep{ding2023cceval} then parse and extract the cross-file dependencies using \coder. During retrieval, the abstracts and the snippets are placed together in the candidate pool (Sec. \ref{sec:candidate_pool}). Note that we keep the cursor positions and the expected outputs the same as those in the original CrossCodeEval (See Appendix~\ref{app:datasets} for more details).

\subsection{Experimental Setup}
\label{subsec:exp_setup}
\noindent\textbf{In-file Context Only.}
We only provide the original code file, where the cursor position is located. Without any explicit cross-file context as input, the model should rely on its knowledge-based reasoning capabilities to accomplish the code generation process.

\noindent\textbf{Retrieval in CrossCodeEval.} This setup is similar to the ``\revision{+ Retrieval}'' setting of CrossCodeEval \citep{ding2023cceval}, where the retrieval candidates are constructed by exhaustively scanning files in the same repository and extracting contiguous-$M$-line code fragments ($M=10$). Then, the top 5 similar candidates and the test query assemble the prompt. Note that for better baseline performance, we do not truncate cross-file contents into a total length of 512 tokens as \citep{ding2023cceval}. 

\noindent\textbf{Retrieval with \coder.}
Based on \coder in Sec. \ref{sec:enhance}, given a repository, we first construct the candidate retrieval pool via abstract and snippet context extraction. Then, we build the retrieval query including prefix length $P=5$ and suffix length $S=5$. After that, candidates are concatenated to the beginning of the in-file context based on their Jaccard similarity \citep{jaccard1912distribution} in descending order, until the cumulative length, including the in-file context, reaches the maximum token limit of $N=4096$.  Note that we set the number of the abstract context $K$ as 3. 

\noindent\textbf{\Finetune.}
To further boost the performance, we fine-tune code LLMs on the training split of \benchmark including 400,000 files mentioned in Sec. \ref{sec:data_generation}. 

\noindent\textbf{Evaluation Metrics.}
Following CrossCodeEval \citep{ding2023cceval}, we evaluate code LLMs with code match metrics including exact match (EM) and edit similarity (ES). Exact match expects that the prediction and the real code are extremely identical, while edit similarity flexibly allows the generated samples to match the original code through a certain number of editing operations.

\subsection{Results}
We present results on \benchmark in Table~\ref{tab:main_benchmark}, CrossCodeEval \citep{ding2023cceval} in Table~\ref{tab:main_cceval}, and \newcceval in Table~\ref{tab:main_cceval+}. 
Based on Table~\ref{tab:main_cceval} and Table~\ref{tab:main_cceval+}, cross-file context is undoubtedly efficacious, while retrieval with \coder surely exhibits its superiority compared with retrieval method in CrossCodeEval. Moreover, fine-tuning strategy significantly improves the performance on both benchmarks. Besides, when abstract contexts are removed from the candidate retrieval pool, these code LLMs suffer from degraded performance. Notably, in Table~\ref{tab:main_benchmark}, after applying \finetune, the Code Llama-7B, which is the worst in the in-file setting, outperforms the StarCoder-7B. 
Overall, we could say that there exists a performance promotion path, ``In-file'' $\rightarrow$ ``+ \coder w.o. Abs.'' $\rightarrow$ ``+ \coder'' $\rightarrow$ ``+ \Finetune''.
% We can also observe the aforementioned phenomenon in \benchmark, which can be seen from Table~\ref{tab:main_benchmark} as well.

% \jh{Among these code LLMs, Code Llama-7B \citep{roziere2023codellama} performs the worst on average, a probable reason is that Code Llama is prone to generate longer predictions than the references (see Table 3 here \footnote{\href{https://github.com/aixcoder-plugin/aiXcoder-7B}{https://github.com/aixcoder-plugin/aiXcoder-7B}}). Long outputs naturally lead to declines in both EM and ES, with EM suffering more severely.}

\begin{table*}[!htb]
	\centering
	\resizebox{0.98\textwidth}{!}{
		\begin{tabular}{l c c c c c c c c c c}
			\toprule
			\multirow{2}{*}{\bf Model}  &
			\multicolumn{2}{c}{\textbf{Python }}&
			\multicolumn{2}{c}{\textbf{Java }}&
			\multicolumn{2}{c}{\textbf{TypeScript }}&
			\multicolumn{2}{c}{\textbf{C\# }}&
			\multicolumn{2}{c}{\textbf{Average}}\\
			\cmidrule(lr){2-3} \cmidrule(lr){4-5} \cmidrule(lr){6-7} \cmidrule(lr){8-9}\cmidrule(lr){10-11}
			&{EM} &{ES} & {EM} & {ES} & {EM} &{ES} & {EM} & {ES} &{EM} &{ES} \\\midrule
			
			\revision{StarCoder-7B} &
			24.3 & 51.3 & 27.7 & 61.4 & 30.8 & 67.1 & 44.7 & 69.8 & 30.9 & 62.1  \\
			\hdashline
			\noalign{\vskip 0.4ex}\; + \revision{Retrieval}&
			27.0 & 51.0 & 33.6 & 62.9 & 33.1 & 66.7 & 47.1 & 67.7 & 34.1 & 61.9  \\
			\hdashline
			\noalign{\vskip 0.4ex}\; + \revision{\Finetune}&
			34.2 & 64.3 & 40.2 & 73.7 & 37.8 & 72.4 & 50.6 & 77.8 & 39.6 & 71.5  \\
			\midrule
			
			\revision{Code Llama-7B} & 
			22.0 & 46.8 & 28.6 & 60.4 & 10.1 & 63.0 & 44.5 & 71.3 & 23.4 & 59.6 \\
			\hdashline
			\noalign{\vskip 0.4ex}\; + \revision{Retrieval} & 
			23.0 & 45.3 & 33.8 & 61.3 & 10.7 & 63.1 & 49.3 & 70.1 & 25.8 & 59.2 \\
			\hdashline
			\noalign{\vskip 0.4ex}\; + \revision{\Finetune}& 
			34.7 & 65.4 & 41.0 & 72.7 & 11.4 & 69.5 & 53.2 & 77.8 & 31.5 & 70.6 \\
			\midrule
			\revision{DeepSeekCoder-6.7B} & 
			25.9 & 51.8 & 31.3 & 62.5 & 33.8 & 67.5 & 42.9 & 67.6 & 32.8 & 62.2  \\
			\hdashline
			\noalign{\vskip 0.4ex}\; + \revision{Retrieval} & 
			28.0 & 52.6 & 36.2 & 62.3 & 35.6 & 66.7 & 48.4 & 67.7 & 35.9 & 62.1 \\
			\hdashline
			\noalign{\vskip 0.4ex}\; + \revision{\Finetune}& 
			37.7 & 67.3 & 45.1 & 75.3 & 42.0 & 75.7 & 59.3 & 81.8 & 44.6 & 74.5 \\
			\bottomrule
		\end{tabular}
	}
	\caption{ 
 Exact match (\%) and edit similarity (\%) performance on CrossCodeEval.}
	\label{tab:main_cceval}
\end{table*}

\begin{table*}[!htb]
	\centering
	\resizebox{0.98\textwidth}{!}{
		\begin{tabular}{l c c c c c c c c c c}
			\toprule
			\multirow{2}{*}{\bf Model}  &
			\multicolumn{2}{c}{\textbf{Python }}&
			\multicolumn{2}{c}{\textbf{Java }}&
			\multicolumn{2}{c}{\textbf{TypeScript }}&
			\multicolumn{2}{c}{\textbf{C\# }}&
			\multicolumn{2}{c}{\textbf{Average}}\\
			\cmidrule(lr){2-3} \cmidrule(lr){4-5} \cmidrule(lr){6-7} \cmidrule(lr){8-9}\cmidrule(lr){10-11}
			&{EM} &{ES} & {EM} & {ES} & {EM} &{ES} & {EM} & {ES} &{EM} &{ES} \\\midrule
			
			\revision{StarCoder-7B} &
			24.3 & 51.3 & 27.7 & 61.4 & 30.8 & 67.1 & 44.7 & 69.8 & 30.9 & 62.1 \\
			\hdashline
			\noalign{\vskip 0.4ex}\; + \revision{\coder}&  
            30.9 & 51.9 & 38.1 & 63.6 & 35.5 & 67.7 & 68.4 & 78.2 & 40.7 & 64.5 \\
			\hdashline
			\noalign{\vskip 0.4ex}\; + \revision{\coder w.o. Abs.}&
			30.9 & 51.8 & 35.3 & 62.6 & 34.5 & 67.6 & 65.2 & 77.1 & 39.2 & 64.0 \\
			\hdashline
			\noalign{\vskip 0.4ex}\; + \revision{\Finetune}&
			41.8 & 68.5 & 46.2 & 76.0 & 40.7 & 74.4 & 69.3 & 86.5 & 47.3 & 75.3  \\
			\midrule
			\revision{Code Llama-7B} & 
			22.0 & 46.8 & 28.6 & 60.4 & 10.1 & 63.0 & 44.5 & 71.3 & 23.4 & 59.6 \\
			\hdashline
			\noalign{\vskip 0.4ex}\; + \revision{\coder} & 
			23.6 & 42.9 & 35.6 & 58.5 & 10.6 & 61.6 & 67.8 & 78.5 & 29.7 & 58.9  \\
			\hdashline
			\noalign{\vskip 0.4ex}\; + \revision{\coder w.o. Abs.}&
            23.9 & 43.0 & 34.1 & 58.9 & 11.0 & 62.3 & 65.4 & 78.0 & 29.1 & 59.2 \\
			\hdashline
			\noalign{\vskip 0.4ex}\; + \revision{\Finetune}& 
			43.0 & 71.8 & 48.8 & 76.8 & 13.0 & 73.2 & 70.3 & 87.0 & 39.0 & 76.1 \\
			\midrule
			\revision{DeepSeekCoder-6.7B} & 
			25.9 & 51.8 & 31.3 & 62.5 & 33.8 & 67.5 & 42.9 & 67.6 & 32.8 & 62.2  \\
			\hdashline
			\noalign{\vskip 0.4ex}\; + \revision{\coder} & 
            32.7 & 54.0 & 41.6 & 64.6 & 38.8 & 67.9 & 69.0 & 78.8 & 43.1 & 65.4  \\
			\hdashline
			\noalign{\vskip 0.4ex}\; + \revision{\coder w.o. Abs.}&
			32.5 & 53.7 & 38.8 & 63.5 & 37.8 & 67.5 & 65.7 & 77.5 & 41.5 & 64.7 \\
			\hdashline
			\noalign{\vskip 0.4ex}\; + \revision{\Finetune}& 
			46.9 & 73.5 & 53.2 & 79.3 & 45.8 & 78.2 & 78.9 & 89.9 & 53.6 & 79.3 \\
			\bottomrule
		\end{tabular}
	}
	\caption{ 
 Exact match (\%) and edit similarity (\%) performance on \newcceval.}
	\label{tab:main_cceval+}
\end{table*}

\begin{table*}[!htb]
	\centering
	\resizebox{0.98\textwidth}{!}{
		\begin{tabular}{l c c c c c c c c c c c c}
			\toprule
			\multirow{2}{*}{\bf Ablation Models}  &
			\multirow{2}{*}{\bf $P$} &
			\multirow{2}{*}{\bf $S$} &
			\multicolumn{2}{c}{\textbf{Python }}&
			\multicolumn{2}{c}{\textbf{Java }}&
			\multicolumn{2}{c}{\textbf{TypeScript}}&
			\multicolumn{2}{c}{\textbf{C\# }}&
			\multicolumn{2}{c}{\textbf{Average}}\\
			\cmidrule(lr){4-5} \cmidrule(lr){6-7} \cmidrule(lr){8-9} \cmidrule(lr){10-11}\cmidrule(lr){12-13}
			& & & {EM} &{ES} & {EM} & {ES} & {EM} &{ES} & {EM} & {ES} &{EM} &{ES} \\
			\midrule
			\revision{StarCoder-7B + \Finetune} 
			& 5 & 5&
			\bf 40.8 & \bf 68.5 & \bf 42.0 & \bf 74.1 & \bf 35.6 & \bf68.1 & \bf 50.0 & \bf 76.6 & \bf 42.1 & \bf 71.8  \\
			\midrule
			\noalign{\vskip 0.4ex}\qquad \multirow{2}{*}{Prefix \& Suffix}
			& 5 & 1 &
			38.0 & 64.0 & 41.6 & 71.8 & 31.6 & 64.7 & 48.0 & 74.2 & 39.8 & 68.7 \\
			& 5 & 3 &
			40.0 & 66.8 & 41.6 & 73.8 & 32.0 & 65.5 & 49.2 & 75.6 & 40.7 & 70.4 \\
			\hdashline
			\noalign{\vskip 0.4ex}\qquad Prefix Only 
			& 10 & - 
			& 39.2 & 68.7 & 39.2 & 71.0 & 32.4 & 67.5 & 47.6 & 74.5 & 39.6 & 70.4 \\
			\midrule
			\noalign{\vskip 0.4ex}\qquad \multirow{4}{*}{w.o. Snippet Context} 
			& 1 & 1 &
			24.8 & 59.1 & 36.4 & 71.5 & 26.8 & 64.4 & 36.8 & 70.3 & 31.2 & 66.3 \\
			& 3 & 3 &
			26.4 & 59.7 & 36.0 & 71.5 & 25.6 & 64.0 & 35.6 & 70.0 & 30.9 & 66.3 \\
			& 5 & 5 &
			25.6 & 59.0 & 37.6 & 71.4 & 24.8 & 64.0 & 38.4 & 70.8 & 31.6 & 66.3 \\
			& 10 & 10 &
			24.0 & 58.1 & 36.4 & 71.5 & 25.2 & 64.2 & 35.2 & 68.4 & 30.2 & 65.5 \\
			\hdashline
			\noalign{\vskip 0.4ex}\qquad \multirow{4}{*}{w.o. Abstract Context} 
			& 1 & 1 &
			38.0 & 65.9 & 37.6 & 69.3 & 31.6 & 65.3 & 42.8 & 71.4 & 37.5 & 68.0 \\
			& 3 & 3 &
			37.2 & 62.9 & 39.2 & 68.3 & 31.6 & 62.5 & 42.4 & 69.3 & 37.6 & 65.7 \\
			& 5 & 5 &
			37.2 & 65.1 & 38.8 & 69.8 & 33.2 & 65.2 & 45.2 & 73.3 & 38.6 & 68.4 \\
			& 10 & 10 &
			37.6 & 66.1 & 37.2 & 68.2 & 32.4 & 64.8 & 44.8 & 72.8 & 38.0 & 68.0 \\
			\bottomrule
		\end{tabular}
	}
			\caption{ 
   Exact match (\%) and edit similarity (\%) performance on \benchmark for StarCoder-7B.
   % measured by exact match (\%) and edit similarity (\%) on the validation set of \benchmark. ``+ \coder w/ R.F.T.'' indicates that in addition to introducing \coder at test time, we further fine-tune the code model on the training set of \benchmark, within \coder as the retriever.
   During retrieval, $P$ prefix lines and $S$ suffix lines are concatenated as the query. In ``Prefix Only'', no suffix lines are taken as the query. ``w.o.'' stands for removing the snippet contexts or abstract contexts at the fine-tuning and the inference stages simultaneously.}
		\label{tab:ablation}
\end{table*}
\subsection{Analysis}
\noindent\textbf{Ablations.}
We conduct ablation studies on \coder and present the results in  Table~\ref{tab:ablation}. We fine-tune StarCoder-7B \citep{li2023starcoder} with \coder and report the performance on the validation set of \benchmark. Meanwhile, we conduct experiments on the retrieval manner, where we replace the FIM mode with ``prefix-only''. We could find that suffix contexts are necessary as the results of $P=5, S=3$ even surpass those of $P=10$. Additionally, we remove either snippet contexts (``w.o. Snippet Contexts'') or abstract contexts (``w.o. Abstract Contexts'') simultaneously in fine-tuning and testing. Results on various prefix lengths ($P$) and suffix lengths ($S$) are further reported for the ablation model. Despite the larger impact without using snippet contexts, abstract contexts still deliver 3.5\% and 3.4\% performance gains. Furthermore, a case study in Fig.~\ref{fig:case_abstract} shows how the code LLMs benefit from abstract context extraction.

\noindent\textbf{Context Perturbation Helps Cope with Confusing Contexts.}
In real-world scenarios, retrievers would produce low-quality results as shown in Fig.~\ref{fig:case_perturbation}. To enhance the model robustness on irrelevant contexts, we select $Q\%$ of the cursors and randomly remove the top similar contexts during \finetune. Fig.~\ref{fig:perturbation} visualizes how the performance of StarCoder-7B on EM and ES scales when using different perturbation rates for \finetune. The perturbation rate $Q\%$ for fine-tuning varies in $0\%, 10\%, 20\%, 50\%,$ and $80\%$, while the perturbation rate for the validation set is always kept as 10\%. The best results are achieved under perturbation rate $Q\%=10\%$, which is also the default setting of \finetune for Table~\ref{tab:main_benchmark} , \ref{tab:main_cceval} and \ref{tab:main_cceval+}. 

\begin{table*}[!htb]
	\centering
	\resizebox{0.98\textwidth}{!}{
	\begin{tabular}{cccccccccccc}
		\toprule
		\multirow{2}{*}{\textbf{Similarity Metric}} & 
		\multicolumn{2}{c}{\textbf{Python}} & 
		\multicolumn{2}{c}{\textbf{Java}} & 
		\multicolumn{2}{c}{\textbf{TypeScript}} & 
		\multicolumn{2}{c}{\textbf{C\#}} & 
		\multicolumn{2}{c}{\textbf{Average}} & 
		\multirowcell{2}{\bf Execution Time \\ (ms)} \\
		\cmidrule(lr){2-3} \cmidrule(lr){4-5} \cmidrule(lr){6-7} \cmidrule(lr){8-9}\cmidrule(lr){10-11}
		& EM  & ES & EM & ES & EM & ES & EM & ES & EM & ES &  \\
		\midrule
		\noalign{\vskip 0.4ex}\;Jaccard &
		\bf 40.8 & \bf 68.5 & \bf 42.0 & \bf 74.1 & \bf 35.6 & \bf 68.1 & \bf 50.0 & \bf 76.6 & \bf 42.1 & \bf 71.8 & \bf 2.941 \\
		\hdashline
		\noalign{\vskip 0.4ex}\;BM25 &
		38.4 & 64.6 & 40.8 & 71.5 & 31.6 & 65.6 & 50.0 & 75.4 & 40.2 & 69.3 & 3.142 \\
		\hdashline
		\noalign{\vskip 0.4ex}\;UniXCoder &
		39.2 & 66.0 & 40.8 & 70.9 & 33.6 & 66.1 & 49.2 & 75.9 & 40.7 & 69.7 & 250.9 \\
		\bottomrule                                                                
	\end{tabular}
	}
	\caption{Performance of StarCoder-7B with different similarity metrics on the validation set of \benchmark. The retrieval time consumption (ms) per sample is further reported.}
	\label{tab:retriever}
\end{table*}

\begin{wrapfigure}{r}{0.35\textwidth}
	\centering
 \vspace{-2mm}
	\includegraphics[width=0.35\textwidth]{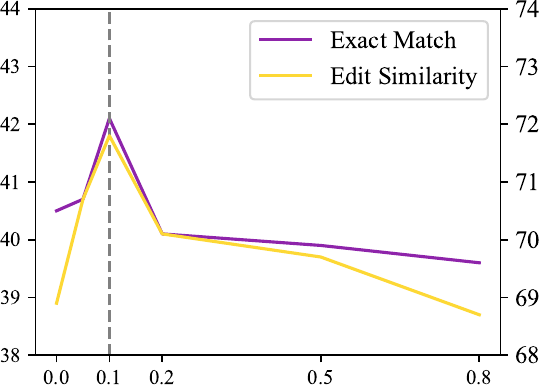}
 \vspace{-3mm}
	\caption{Performance of StarCoder-7B + \Finetune at various perturbation rates on the validation set of \benchmark.}
	\label{fig:perturbation}
 \vspace{-8pt}
\end{wrapfigure} 

\noindent\textbf{Lexical Retrievers are Cost-effective.}
For ranking-based code retrieval, several approaches could be adopted as the similarity metric. We categorize the metrics into lexical ones, including Jaccard similarity \citep{jaccard1912distribution} and BM25 \citep{robertson2009probabilistic}, and neural ones such as UniXcoder \citep{guo2022unixcoder}. In real-world scenarios, we often need to make a trade-off, as lexical retrievers could save more response time (RT) while neural retrievers would theoretically achieve higher performance. To find the most suitable retriever for \coder, we evaluate the code match performance and retrieval duration with Jaccard similarity, BM25, and UniXCoder. Table~\ref{tab:retriever}  presents the code match performance and retrieval duration of each retriever. Although UniXCoder outperforms BM25 on code match, its time consumption is extremely worse than lexical retrievers. We find that Jaccard similarity is the best retriever regarding both effectiveness and efficiency. Thus, we choose Jaccard similarity by default. See Appendix~\ref{app:time} for details of execution time benchmarking.
% on repo-level code completion tasks.

\noindent\textbf{Scalability of Model Performance.}
We present how the performance of StarCoder scales w.r.t model parameters in Table~\ref{tab:abs_modelscale}. We find that almost no performance gain is obtained for StarCoder-3B when using \coder with abstract and snippet contexts.
% struggles when directly inferencing given abstract context in \coder. 
After \finetune, StarCoder-3B can outperform the inference-only results of StarCoder-7B. In Table~\ref{tab: abs_datascale} we further evaluate the fine-tuned StarCoder-7B when using different sizes of tuning datasets based on our \finetune.
We observe that when increasing from 10K samples to 100K samples per language,
better results are obtained.
When continually increasing,
the results are relatively stable.
Thus,
we choose 10K samples per language as the default training set size of \benchmark.

\noindent\textbf{Scalability on Multi-line Predictions.}
In Fig.~\ref{fig:multiline},
we provide a detailed analysis of the scalability on multi-line predictions.
Specifically,
in Fig.~\ref{fig:multiline}(a),
we provide the counts of completion references with different lines and observe that our \benchmark has more multi-line completion scenes.
Then,
in Fig.~\ref{fig:multiline}(b) and Fig.~\ref{fig:multiline}(c),
we provide the results of StarCoder-7B with \Finetune on 
\newcceval and \benchmark,
and observe that the scores on \benchmark are lower than \newcceval a lot, specifically in multi-line scenes,
which means that our \benchmark is more challenging than the \newcceval.

\noindent\textbf{Limitations of CrossCodeEval.}
\label{sec:compare}
We summarize three major drawbacks of CrossCodeEval~\citep{ding2023cceval} as follows:  1) The retrieval results provided in CrossCodeEval are produced by sliding-window-based searching, which could potentially miss long-term dependencies that exceed the length of $M$. In contrast, abstract extraction expands the reception field of code LLMs with coarse-grained global information. 2) CrossCodeEval only considers explicit dependencies (i.e., file imports), but there could be implicit references during software development. For example, in Java, usages in the same package do not require the \texttt{import} statement. As shown in Fig.~\ref{fig:implicit}, such implicit dependencies could not be detected by \citet{ding2023cceval}, but are successfully covered by our \coder during constructing \newcceval. 3) Compared with CrossCodeEval, our \benchmark provides more complex cross-file dependencies (see ``Avg. \# cross files in prompt'' in Table~\ref{tab:data_stat} and Table~\ref{app:cceval}) and more difficult multi-line test samples in Fig.~\ref{fig:multiline}).

\begin{figure*}[!htb]
	\centering
	\subfigure[Multi-line Reference Counts]{\includegraphics[width=0.3\textwidth]{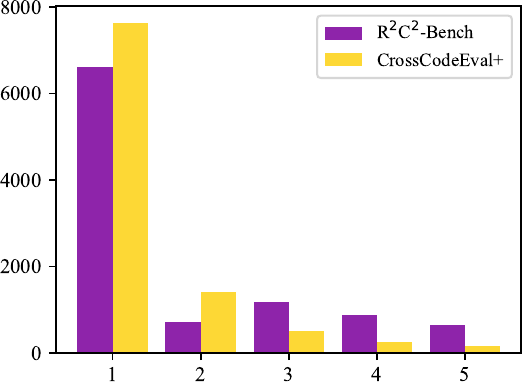}}
	\subfigure[Exact Match]{\includegraphics[width=0.3\textwidth]{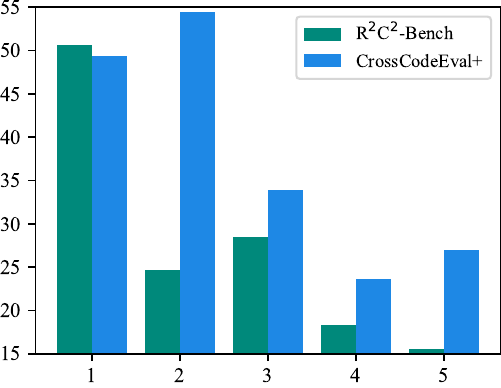}}
	\subfigure[Edit Similarity]{\includegraphics[width=0.3\textwidth]{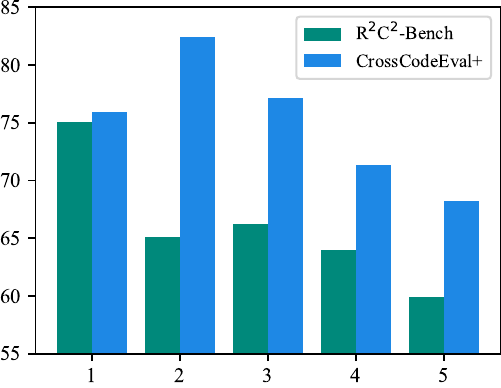}}
	\caption{(a) Statistics of references from 1 to 5 lines in \benchmark and \newcceval. (b) Exact Match of StarCoder-7B w/ \finetune on \benchmark and \newcceval when expected output varies from 1 to 5 lines. (c) Edit Similarity of StarCoder-7B w/ \finetune on \benchmark and \newcceval when expected output varies from 1 to 5 lines.}
	\label{fig:multiline}
\end{figure*} 
\begin{table*}[!htb]
	\centering
	\resizebox{0.98\textwidth}{!}{
		\begin{tabular}{l c c c c c c c c c c}
			\toprule
			\multirow{2}{*}{\bf Model}  &
			\multicolumn{2}{c}{\textbf{Python }}&
			\multicolumn{2}{c}{\textbf{Java }}&
			\multicolumn{2}{c}{\textbf{TypeScript }}&
			\multicolumn{2}{c}{\textbf{C\# }}&
			\multicolumn{2}{c}{\textbf{Average}}\\
			\cmidrule(lr){2-3} \cmidrule(lr){4-5} \cmidrule(lr){6-7} \cmidrule(lr){8-9}\cmidrule(lr){10-11}
			&{EM} &{ES} & {EM} & {ES} & {EM} &{ES} & {EM} & {ES} &{EM} &{ES} \\\midrule
			
			\revision{StarCoder-3B} &
			10.8 & 38.6 & 18.0 & 51.1 & 12.4 & 49.0 & 14.8 & 50.0 & 14.0 & 47.2  \\
			\hdashline
			\noalign{\vskip 0.4ex}\; + \revision{\coder}&
			16.8 & 34.1 & 18.0 & 39.7 & 16.0 & 39.0 & 12.8 & 31.2 & 15.9 & 36.0  \\
			\hdashline
			\noalign{\vskip 0.4ex}\; + \revision{\Finetune}&
			40.0 & 64.4 & 37.2 & 68.0 & 28.0 & 61.8 & 45.6 & 71.0 & 37.7 & 66.3  \\
			\midrule
			\revision{StarCoder-7B} & 
			15.2 & 45.2 & 26.0 & 59.1 & 14.4 & 52.4 & 16.0 & 52.6 & 17.9 & 52.3  \\
			\hdashline
			\noalign{\vskip 0.4ex}\; + \revision{\coder} & 
			23.2 & 46.4 & 30.8 & 56.8 & 25.2 & 52.6 & 26.4 & 48.9 & 26.4 & 51.2  \\
			\hdashline
			\noalign{\vskip 0.4ex}\; + \revision{\Finetune}& 
			40.8 & 68.5 & 42.0 & 74.1 & 35.6 & 68.1 & 50.0 & 76.6 & 42.1 & 71.8  \\
			\bottomrule
		\end{tabular}
	}
	\caption{Performance on the validation set of \benchmark.}
	\label{tab:abs_modelscale}
\end{table*}
\begin{table*}[!htb]
	\centering
	\resizebox{0.98\textwidth}{!}{
		\begin{tabular}{l c c c c c c c c c c}
			\toprule
			\multirowcell{2}{\bf Training samples \\ ($\times 4$)}  &
			\multicolumn{2}{c}{\textbf{Python }}&
			\multicolumn{2}{c}{\textbf{Java }}&
			\multicolumn{2}{c}{\textbf{TypeScript }}&
			\multicolumn{2}{c}{\textbf{C\# }}&
			\multicolumn{2}{c}{\textbf{Average}}\\
			\cmidrule(lr){2-3} \cmidrule(lr){4-5} \cmidrule(lr){6-7} \cmidrule(lr){8-9}\cmidrule(lr){10-11}
			&{EM} &{ES} & {EM} & {ES} & {EM} &{ES} & {EM} & {ES} &{EM} &{ES} \\
			\midrule
			\noalign{\vskip 0.4ex}\;\qquad\revision{10,000} &
			38.0 & 63.7 & 40.4 & 71.3 & 31.2 & 65.6 & 44.0 & 74.4 & 38.4 & 68.7    \\
			\hdashline
			\noalign{\vskip 0.4ex}\;\qquad\revision{50,000}&
			40.0 & 67.2 & 39.6 & 71.1 & 34.0 & 66.3 & 49.2 & 75.6 & 40.7 & 70.0    \\
			\hdashline
			\noalign{\vskip 0.4ex}\;\qquad\revision{100,000}&
			\bf 40.8 & \bf 68.5 & 42.0 & \bf 74.1 & \bf 35.6 & \bf 68.1 & \bf 50.0 & \bf 76.6 & \bf 42.1 & \bf 71.8    \\
			\hdashline
			\noalign{\vskip 0.4ex}\;\qquad\revision{200,000}&
			39.6 & 66.9 & \bf 43.6 & 73.9 & 35.4 & 66.0 & 49.6 & 75.6 & 42.1 & 70.6 \\
			\bottomrule
		\end{tabular}
	}
	\caption{Performance of StarCoder-7B with \Finetune on the validation set of \benchmark. Experiments are conducted on \benchmark using different sizes of training data.}
	\label{tab: abs_datascale}
\end{table*}

\section{Conclusion}
In this paper,
we propose the R$^2$C$^2$-Coder including R$^2$C$^2$-Enhance and R$^2$C$^2$-Bench to enhance and benchmark the real-world repository-level code completion abilities of code LLMs.
Specifically, first, R$^2$C$^2$-Enhance is a code prompt construction method,
which includes candidate retrieval pool construction and completion prompt construction.
Second,
the R$^2$C$^2$-Bench is a more challenging repository-level code completion dataset with R$^2$C$^2$-Bench with training, validation, and test splits.
Comprehensive results on multiple benchmarks demonstrate the effectiveness of our R$^2$C$^2$-Coder.

\section{Acknowledgement}
This research project is supported by the \textbf{Alibaba Aone Copilot Group}.
\bibliography{custom}
\bibliographystyle{elsarticle-harv}
\appendix
\begin{figure}[!htb]
	\centering
	\includegraphics[width=0.98\textwidth]{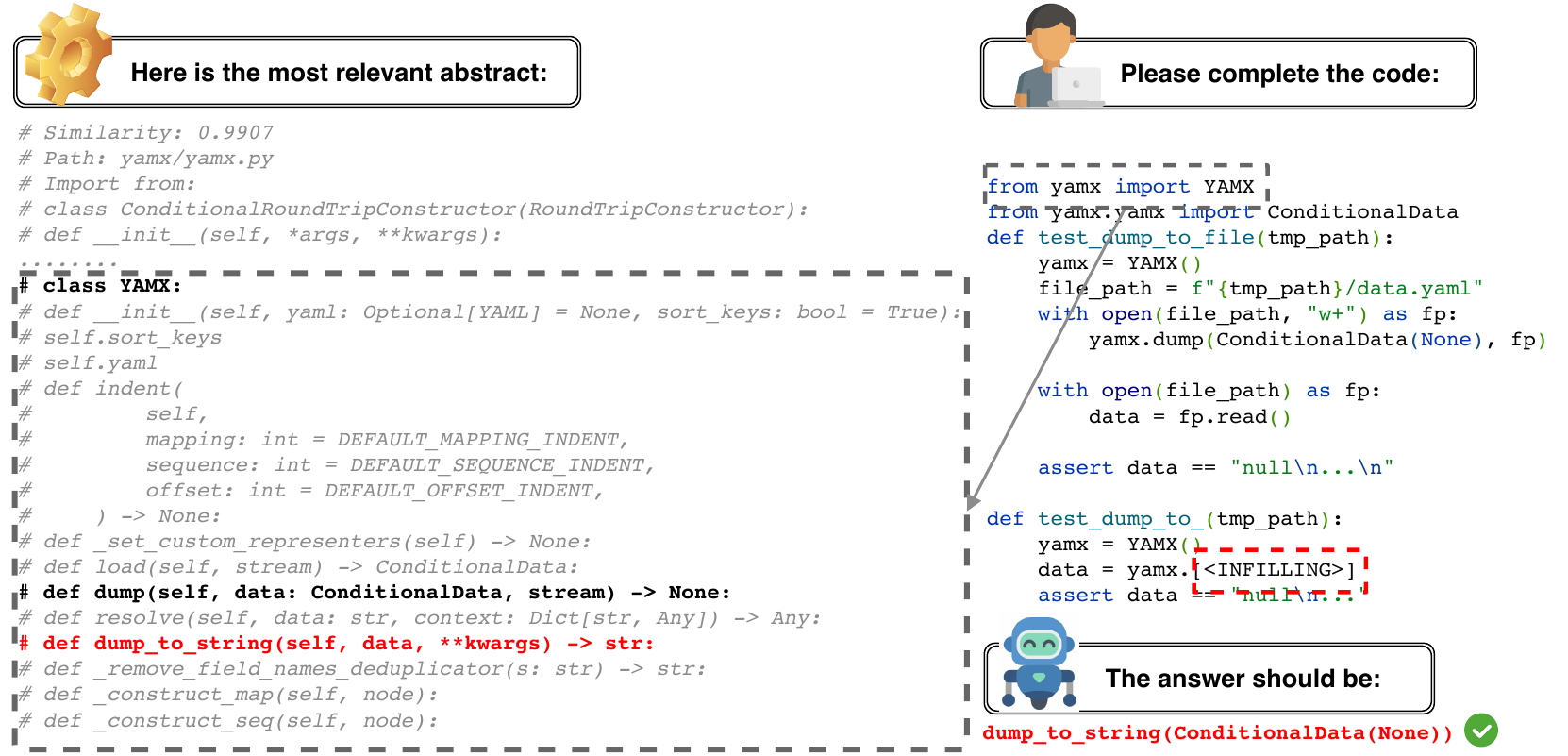}
	\caption{A test case from \newcceval. With the help of abstract extraction, the code model could understand the entire class \texttt{YAMX} at a glance and then generate the correct method invocation.}
	\label{fig:case_abstract}
\end{figure}
\begin{figure}[!htb]
	\centering
	\includegraphics[width=0.98\textwidth]{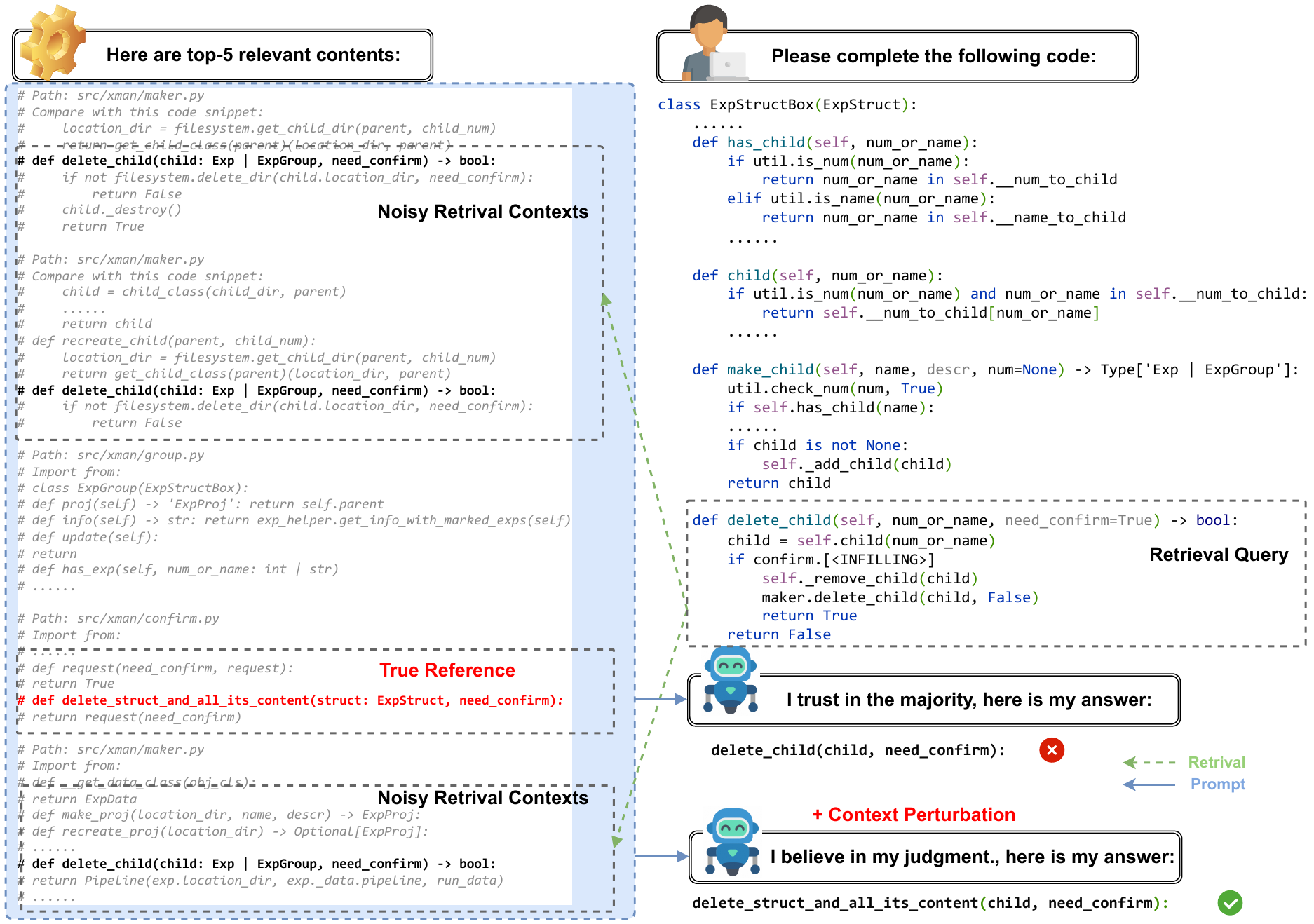}
	\caption{A test case from \benchmark. The statement \texttt{delete\_child} occurs multiple times in the prefix and suffix, resulting in retrieval contexts that are highly relevant to \texttt{delete\_child}. The vanilla code model tends to predict the term that frequently appears in retrieval results. On the contrary, the model could focus on the true reference after \finetune.}
	\label{fig:case_perturbation}
\end{figure}
\section{Broader Impacts and Limitations}
\label{app:limitation}
\noindent\textbf{Broader Impacts.} {In this paper, we propose a methodology with prominent efficiency and ingenious utilization of cross-file information for real-world code completion. We anticipate that this methodology, alongside our fine-tuning strategy and the corresponding benchmarks, will assist developers in enhancing coding efficiency and liberating productivity. Therefore, we hope our work can enable programmers to focus on scenarios with greater creativity.}

\noindent\textbf{Limitations.} Firstly, to attain enhanced performance, it is essential to fine-tune several hyperparameters. Secondly, despite the need for only 400,000 training samples to achieve outstanding performance, the process demands a certain amount of GPU resources for fine-tuning. Thirdly, four programming languages are covered in our setting, and a large number of languages are not considered in our paper.
In our future work, we will continue to investigate to support more languages for facilitating research in this field. 

\section{More Details}
\subsection{Details on Abstract Context Extraction}
\label{app:ast_sec}
In Fig.~\ref{app:ast},
we have provided a relatively complex abstract syntax tree for the code in Fig.~\ref{app:code}.
Besides,
we also provide two examples of the extracted abstract context for Python and Java in Fig.~\ref{fig:abs}.
\begin{figure}[t]
    \centering
    \includegraphics[width=0.8\linewidth]{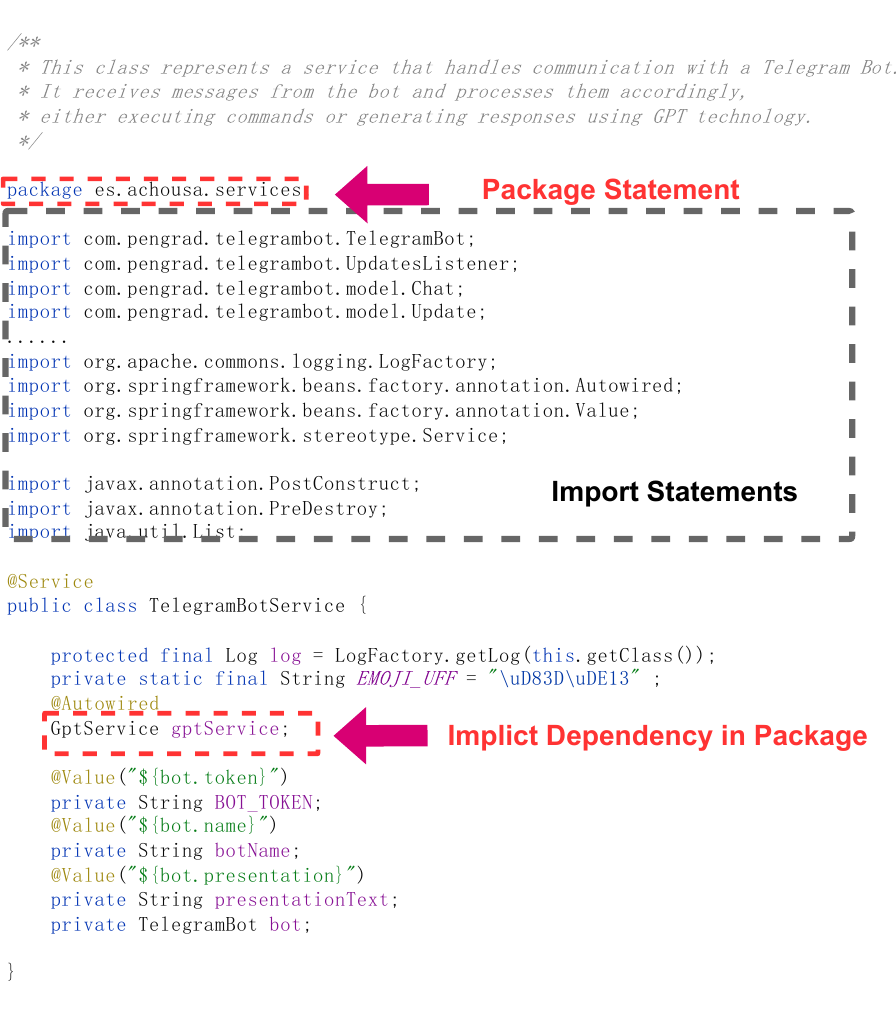}
    \caption{A case found in CrossCodeEval \citep{ding2023cceval}. As Java supports the \texttt{package} statement, developers could call cross-file APIs in the same package. Such kinds of implicit dependencies are ignored by \citet{ding2023cceval}.}
    \label{fig:implicit}
\end{figure} 
% \begin{wrapfigure}{r}{0.4\textwidth}
% 	\centering
%     \vspace{-2mm}
% 	\includegraphics[width=0.4\textwidth]{implicit.pdf}
%     \vspace{-2mm}	
%  \caption{A case found in CrossCodeEval \citep{ding2023cceval}. As Java supports the \texttt{package} statement, developers could call cross-file APIs in the same package. Such kind of implicit dependencies are ignored by \citet{ding2023cceval}.}
% 	\label{fig:implicit}
%  \vspace{-2mm}
% \end{wrapfigure}
\begin{figure}[t]
    \centering
    \includegraphics[width=1.0\linewidth]{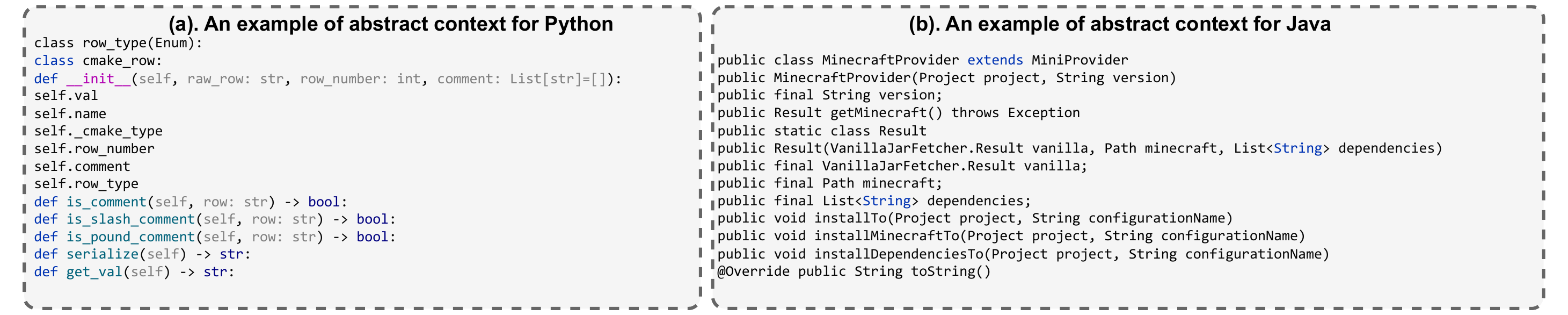}
    \caption{Examples of the generated abstract contexts for Python and Java.}
    \label{fig:abs}
\end{figure} 
\begin{figure}[t]
    \centering
    \includegraphics[width=0.25\linewidth]{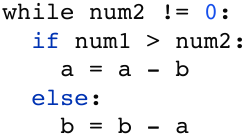}
    \caption{A code example.}
    \label{app:code}
\end{figure} 

\begin{figure}[t]
    \centering
    \includegraphics[width=1.0\linewidth]{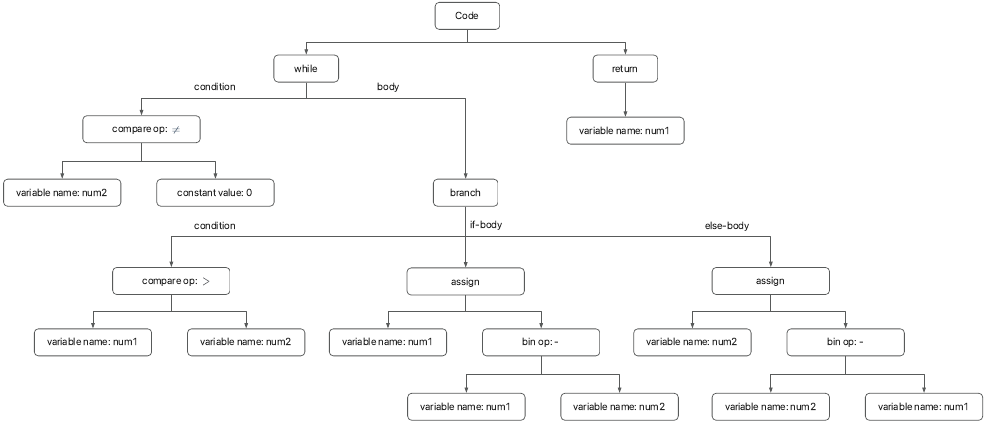}
    \caption{An example of the abstract syntax tree generated by the Tree-sitter tool.}
    \label{app:ast}
\end{figure} 
\subsection{Quality Control Procedure}
\label{app:quality}
% \begin{figure}[t]
%     \centering
%     \includegraphics[width=1.0\linewidth]{Styles/Appendix-AST.pdf}
%     \caption{An example of the abstract syntax tree generated by the Tree-sitter tool.}
%     \label{app:ast}
% \end{figure} 
% \subsection{Quality Control}
% \label{app:quality}
We begin by establishing the ground-truth completion result as the ``reference'' and create a set of post-processing filters (i.e., rule-based and model-based rules) to enhance dataset quality.
Initially, we eliminate examples based on two criteria: (1) references that are too short (fewer than 10 tokens) and (2) those that are overly long (exceeding 5 lines). Next, we ensure that more than 30\% of our references are in multiple lines.
To prevent predictable inference from the current file—potentially due to prominent indicators in function names and comments—we input the examples (input prompts) into the \texttt{DeepSeekCoder-1.3B} model~\citep{guo2024deepseek} for completion and discard any exact matches. This step results in the removal of 57.1\%, 39.8\%, 48.0\%, and 47.2\% of samples for Java, Python, TypeScript, and C\#, respectively.
Lastly, we conducted human annotations on a subset of \benchmark and observed that the dataset possesses sufficient quality for cross-file code completion purposes.

% We first define the ground-truth completion result as the ``reference'', and design a series of rule-based and model-based post-processing filters to ensure the quality of the dataset as follows. 
% First,
% we filter examples if (1) too short ($<10$ tokens) reference. (2). too long ($5$ lines) reference.
% Second,
% we keep that the percentage of the multiple lines for the reference is larger than 30\%.
% Moreover, to ensure that the reference isn't predictably inferred solely from the current file (possibly owing to strong clues in function names and comments), we feed the examples (input prompts) to \texttt{DeepSeekCoder-1.3B} model~\citep{guo2024deepseek} to complete the statement and remove the exact matches. This step removes 57.1\%, 39.8\%, 48.0\%, and 47.2\% of the samples for Java, Python, TypeScript, and C\#, respectively. 
% Finally, we perform human annotations on a subsample of \benchmark and found that the dataset has a satisfactory quality to serve the goal of cross-file code completion. 

\subsection{Dataset Statistics of the validation split of \benchmark}
\label{app:val_r2coder}
In Table~\ref{tab:val_r2coder},
we provide the statistics of the validation split of \benchmark.

\begin{table}
% \vspace{-8pt}
\centering
% \def\arraystretch{1.0}%
% \resizebox{1\linewidth}{!}
% {
\begin{tabular}{l r r r r}
\toprule
Feature & Python & Java & TypeScript & C\#\\
\midrule
\# Repositories & 23 & 43 & 42 & 21 \\
\# Files & 250 & 250 & 250 & 250 \\
\# Examples & 425 & 717 & 687 & 521 \\
Avg. \# cross files in prompt  &6.03&7.76&6.84&11.28\\
Avg. \# lines in prompt & 326.04 & 333.98 & 367.46 & 374.81 \\
Avg. \# tokens in prompt & 3527.72& 3467.47 & 3451.46 & 3460.65\\
Avg. \# lines in reference & 1.65& 1.80 & 2.12 & 1.77 \\
Avg. \# tokens in reference &16.19 &14.64 & 15.06 & 13.52 \\
\bottomrule
\end{tabular}
\caption{The statistics of the validation split of \benchmark.} 
\label{tab:val_r2coder}
% \vspace{-2mm}
\end{table}

\subsection{Dataset Statistics of CrossCodeEval}
\label{app:cceval}
As shown in Table~\ref{tab:data_stat_cceval},
we have provided the dataset statistics of the CrossCodeEval benchmark.
\begin{table}
% \vspace{-8pt}
\centering
% \def\arraystretch{1.0}%
% \resizebox{1\linewidth}{!}
% {
\begin{tabular}{l r r r r}
\toprule
Feature & Python & Java & TypeScript & C\#\\
\midrule
\# Repositories & 471 & 239 & 193 & 99 \\
\# Files & 1368 & 745 & 779 & 642 \\
\# Examples & 2665 & 2139 & 3356 & 1768 \\
Avg. \# cross files in prompt  &3.19&3.15&3.16&3.98\\
Avg. \# lines in prompt & 90.6 & 106.7 & 116.5 & 71.1 \\
Avg. \# tokens in prompt & 938.9 & 995.3 & 944.9 & 584.1 \\
Avg. \# lines in reference & 1.0 & 1.1 & 1.7 & 1.7 \\
Avg. \# tokens in reference & 13.2 & 14.5 & 17.4 & 12.5 \\
\bottomrule
\end{tabular}
\caption{The statistics of the CrossCodeEval.} 
\label{tab:data_stat_cceval}
% \vspace{-2mm}
\end{table}

\section{More Experiments}
\subsection{Details of the baseline models}
\label{app:models}
\noindent\textbf{StarCoder} \citep{li2023starcoder} is a generative decoder-only LM series whose size varies from 1B to 15.5B. Trained on the Stack \citep{kocetkov2022stack} dataset, StarCoder supports a maximum context length of 8K. 
%We also conduct experiments within the recently released StarCoder 2 \citep{lozhkov2024starcoder2}, which is trained on a 4 times larger corpus than its predecessor.

\noindent\textbf{DeepSeekCoder} \citep{guo2024deepseek} is a suite of code models with sizes ranging from 1.3B to 33B trained on a 2-trillion-token manual collected corpus. Benefit from FIM strategy \citep{bavarian2022efficient} and Rotary Position Embedding (RoPE) \citep{su2024roformer}, these models are capable of efficient code generation and infilling under a 16K context window. 

\noindent\textbf{Code Llama} \citep{roziere2023codellama} is a family of large language models based on Llama 2 \citep{touvron2023llama} with 7B, 13B, 34B, and 70B parameters. These models are trained on sequences of 16K tokens and capable of inference on 100K. We conduct experiments with the foundational Code Llama models, not the Python-specialized versions or instruction-tuned ones.

\subsection{Details of CrossCodeEval and \newcceval}
\label{app:datasets}
\noindent\textbf{CrossCodeEval} \citep{ding2023cceval} is an arising benchmark for cross-file code completion that is built on a diverse set of real-world, open-sourced, permissively-licensed Github repositories in four popular programming languages: Python, Java, TypeScript, and C\#. To create accurate and high-quality test samples, the authors adopt static analysis and filter out those samples solved by \texttt{starcoderbase-1B} within in-file only context.

\noindent\textbf{\newcceval.} We collect the original code bases of CrossCodeEval \citep{ding2023cceval} then parse and extract the cross-file dependencies with our \coder. During retrieval, the abstracts and the snippets are placed together in the candidate pool (Sec. \ref{sec:candidate_pool}). Note that we keep the cursor positions and the expected outputs the same as those in CrossCodeEval. Here, we name the reconstructed CrossCodeEval as \newcceval.
\begin{figure*}[!htb]
	\centering
	\subfigure[StarCoder]{\includegraphics[width=0.3\textwidth]{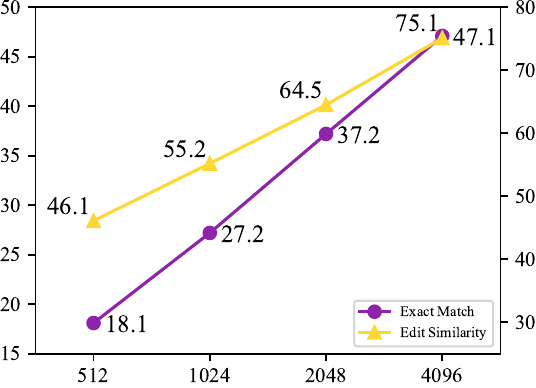}}
	\subfigure[CodeLlama]{\includegraphics[width=0.3\textwidth]{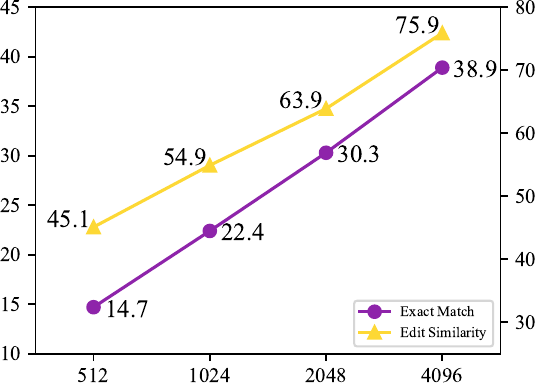}}
	\subfigure[DeepSeekCoder]{\includegraphics[width=0.3\textwidth]{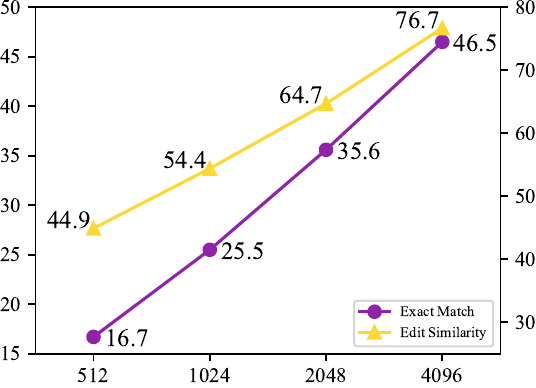}}
	\caption{Performance on CrossCodeEval after \finetune with various input lengths.}
	\label{fig:input}
\end{figure*} 

\subsection{Scalability of Model Performance on Various Input Lengths}
As depicted in Figure~\ref{fig:input}, we report results produced by code models on CrossCodeEval \citep{ding2023cceval} when the input lengths of \finetune range in \{512, 1024, 2048, 4096\} tokens. There exists a scaling law that the performance is improved when input length increases.

\subsection{Execution Time Benchmarking}
\label{app:time}
Execution time measurements are conducted on an Intel Xeon CPU E5-2682 v4 with a base frequency of 2.50 GHz and a turbo boost of up to 4.00 GHz.

% \subsection{Performance with different truncation length}
% \noindent\textbf{\benchmark} is built on permissively licensed repositories from GitHub. Following CrossCodeEval \citep{ding2023cceval}, we select repos mainly written in Python, Java, TypeScript, and C\#. We adopt a suite of filters for quality control and decontamination then utilize tools including \texttt{tree-sitter} for abstract context extraction and cursor position generation. For training the model to cope with potentially low-quality retrieval results, we further perform context perturbation on the training set of \benchmark Sec(\ref{sec:data_generation}). 

\end{document}